\documentclass[11pt]{article}

\usepackage{acl}

\usepackage{times}
\usepackage{latexsym}
\usepackage[T1]{fontenc}
\usepackage[utf8]{inputenc}
\usepackage{microtype}
\usepackage{inconsolata}
\usepackage{graphicx}
\usepackage{amsmath}
\usepackage{amssymb}
\usepackage{booktabs}
\usepackage{multirow}
\usepackage{subcaption}
\usepackage{xcolor}
\usepackage{tikz}
\usepackage{pgfplots}
\pgfplotsset{compat=1.18}
\usepgfplotslibrary{fillbetween}
\usetikzlibrary{shapes.geometric, arrows.meta, positioning, fit, backgrounds, calc, patterns}
\usepackage[ruled,vlined,linesnumbered]{algorithm2e}
\usepackage{enumitem}
\usepackage{orcidlink}

\title{Model Collapse as Cultural Evolution}

\author{
  Dongxin Guo\,\orcidlink{0009-0000-2388-1072} \\
  The University of Hong Kong \\
  Hong Kong, China \\
  \texttt{bettyguo@connect.hku.hk} \\
  \And
  Jikun Wu\,\orcidlink{0000-0002-2327-4157} \\
  Stellaris AI Limited \\
  Hong Kong, China \\
  \texttt{hk950014@connect.hku.hk} \\
  \And
  Siu Ming Yiu\,\orcidlink{0000-0002-3975-8500} \\
  The University of Hong Kong \\
  Hong Kong, China \\
  \texttt{smyiu@cs.hku.hk} \\
}

\begin{document}
\maketitle

\begin{abstract}
	Model collapse, the progressive degradation of LLMs trained on their own outputs, has been characterized statistically but lacks a \emph{linguistic} explanation for which structures degrade, in what order, and why. We show that iterated learning theory from cultural evolution fills this gap. We derive five falsifiable predictions, distinguish those uniquely discriminative for the theory from confirmatory ones, and test them by self-training LLaMA-2-7B and Mistral-7B over 10 generations in English, German, and Turkish. The critical discriminative finding: compositionality follows a \emph{non-monotonic} trajectory (initially rising, then falling) under unfiltered self-training. This signature persists with maximally regular seed data (ruling out noise removal) and is sustained only by task-grounded filtering, not random filtering, providing the first LLM-scale evidence for the compression--communication tradeoff. All predictions are confirmed with large effect sizes (Hedges' $g > 1.6$; $\text{BF}_{10} > 100$), and LLM regularization gradients closely match human behavioral data ($R^2 = 0.94$). These results reframe model collapse as a cultural transmission phenomenon and yield concrete principles for self-training pipeline design.
\end{abstract}

\section{Introduction}
\label{sec:intro}

Why are human languages structured the way they are? Cultural evolution theory offers a compelling answer: languages are shaped by the process of being learned and transmitted \citep{christiansen2008language, christiansen2016now}. When language passes through chains of learners, each generation's inductive biases are amplified, progressively eliminating unpredictable variation and promoting learnable structure \citep{kirby2008cumulative, kirby2014iterated}. This process, formalized as \emph{iterated learning} \citep{griffiths2007language}, provides a mathematical explanation for core properties of language: regularity, compositionality, and Zipfian frequency distributions emerge from the dynamics of cultural transmission itself. Critically, \citet{kirby2015compression} demonstrated that compositionality requires both compression pressure (from the learning bottleneck) and communication pressure (from the need to be expressive); compression alone produces degenerate languages.

In a seemingly related development, training LLMs on their own outputs causes \emph{model collapse}: distributional tails disappear, diversity decreases, and outputs converge toward degenerate forms \citep{smith2024preventing, alemohammad2024selfconsuming}. This phenomenon has been characterized through statistical estimation theory \citep{dohmatob2024tale, dohmatob2024model, dohmatob2025strong}, with \citet{guo2024curious} documenting linguistic diversity decline across several dimensions and \citet{seddik2024model} providing information-theoretic analysis. While these accounts accurately describe the statistical mechanics of collapse, they provide limited insight into \emph{which} linguistic properties are affected, in what order, and why.

The conceptual connection between model collapse and cultural evolution has been established by several recent contributions. \citet{ren2024improving} formally showed that LLM self-evolution instantiates Bayesian iterated learning and proved monotonic bias amplification across generations. \citet{smith2024preventing} argued in a \emph{Nature} correspondence that the compression--expressivity tradeoff from human language transmission explains why self-training degrades. \citet{kouwenhoven2025searching} tested iterated learning with LLMs in referential games, finding that LLMs produce degenerate vocabularies without sufficient compression pressure, while \citet{galke2024what} reviewed language evolution insights for LLMs broadly. We build on these foundations by providing the first systematic test of \emph{discriminative} predictions (those that distinguish iterated learning from generic model collapse) at full LLM scale on natural language.

We derive five predictions from cultural evolution theory and explicitly distinguish their discriminative status:

\paragraph{P1: Frequency-dependent loss order (\emph{partially discriminative}).} Low-frequency variants should be lost before high-frequency ones, following a specific log-linear gradient matching human regularization \citep{reali2009frequency, morgan2016frequency}. Generic collapse predicts tail loss but not the specific functional form.

\paragraph{P2: Monotonic regularity increase (\emph{confirmatory}).} Morphological patterns should regularize across typologically distinct languages, reflecting simplicity bias amplification \citep{shah2020pitfalls, kallini2024mission}.

\paragraph{P3: Dimension-specific degradation rates (\emph{partially discriminative}).} Pragmatic constructions should be lost before syntactic ones, reflecting the formal--functional dissociation in LLMs \citep{mahowald2024dissociating}.

\paragraph{P4: Non-monotonic compositionality (\emph{uniquely discriminative}, critical test).} Without communicative grounding, compositionality should follow a non-monotonic trajectory (initially rising, then falling) even from maximally regular seed data. Only task-grounded evaluation should sustain compositionality \citep{kirby2015compression}. Generic collapse predicts only monotonic degradation.

\paragraph{P5: Distributional narrowing (\emph{confirmatory}).} Zipf exponents should decrease, reflecting tail loss \citep{piantadosi2014zipfs}.

We test all five predictions through controlled experiments with two LLMs, including a \emph{regularized-seed control} and a \emph{three-condition filtering experiment} that isolate the compression--communication mechanism. Morphological regularity is assessed across English, German, and Turkish; construction diversity and compositionality are evaluated in English with collapse-robust metrics. All five predictions are confirmed. The non-monotonic compositionality trajectory (P4) persists with regularized seed data and is sustained only by quality-grounded (not random) filtering, providing the first LLM-scale demonstration of the compression--communication tradeoff. LLM regularization gradients closely match the human curve from \citet{morgan2016frequency}.

Our contributions are: \textbf{(1)}~an explicit extension of the iterated learning framework of \citet{ren2024improving} to natural language, with five falsifiable predictions whose discriminative status is categorized (\S\ref{sec:framework}); \textbf{(2)}~four linguistically grounded metrics with collapse-robust validation (\S\ref{sec:metrics}); \textbf{(3)}~systematic empirical tests including critical controls (regularized-seed, three-condition filter) across 10 generations, two models, and three languages (\S\ref{sec:results}); and \textbf{(4)}~explicit implications for self-training pipeline design (\S\ref{sec:discussion}) and Construction Grammar (Appendix~\ref{app:cxg_discussion}).

\section{Theoretical Framework}
\label{sec:framework}

\subsection{Iterated Learning and the Compression--Communication Tradeoff}

\citet{griffiths2007language} proved that when Bayesian learners observe data generated by a previous learner and produce data for the next, the chain of posterior distributions converges to the learner's prior. \citet{kalish2007iterated} confirmed this empirically: human participants converged on positive linear functions within 1--4 generations regardless of seed conditions, a result critical for our regularized-seed control design (\S\ref{sec:reg_seed}). \citet{reali2009frequency} extended this to frequency-dependent regularization under Beta priors, and \citet{ferdinand2019regularization} demonstrated that these dynamics emerge even with non-Bayesian learners using gradient-based optimization, because the essential mechanism (imperfect learning followed by imperfect reproduction) is shared regardless of the inference procedure.

Three properties are central to our predictions. First, \emph{bottleneck severity}: convergence speed depends on the transmission bottleneck \citep{kirby2014iterated}. Second, the \emph{compression--communication tradeoff}: \citet{kirby2015compression} demonstrated that transmission chains without communicative pressure produce degenerate languages, while chains with communicative pressure produce compositional languages balancing learnability and expressivity. Third, \emph{dimension-specific emergence}: \citet{beckner2017emergence} found that structure emerges along some meaning dimensions before others. Concretely, these three properties map onto our predictions as follows: bottleneck severity drives P1 and P5 (the rate and shape of frequency-dependent tail loss); the compression--communication tradeoff drives P2 and P4 (regularization under compression alone vs.\ the non-monotonic compositionality trajectory that requires communicative pressure to sustain); and dimension-specific emergence drives P3 (the ordering in which constructions are lost). The fine-grained bias--prediction mappings underlying this correspondence are detailed in \S\ref{sec:parallel} and Appendix~\ref{app:bias_mapping}.

\subsection{LLM Self-Training as Iterated Learning}
\label{sec:parallel}

An LLM $\mathcal{M}_n$ trained on data $D_n$ constitutes generation $n$. In self-training, $D_{n+1}$ is sampled from $\mathcal{M}_n$'s output distribution. Each generation implements one step of iterated learning, with the model's inductive biases playing the role of the Bayesian prior.

\citet{ren2024improving} formally established this correspondence, proving that if in-context learning approximates Bayesian inference, the classical convergence result of \citet{griffiths2007language} applies: iterative self-training monotonically amplifies prior biases. Our work extends their framework in two directions. First, we characterize which \emph{specific linguistic phenomena} emerge during this process (non-monotonic compositionality dynamics, frequency-dependent regularization paralleling human acquisition, and dimension-specific construction loss), all invisible under monotonic-amplification models. Second, we test on natural language with weight-based learning (fine-tuning), complementing Ren et al.'s in-context learning experiments.

The mapping between LLM biases and iterated learning priors can be made more precise than a generic ``inductive bias'' characterization. Following \citet{ferdinand2019regularization}'s demonstration that different bias types produce different evolutionary trajectories, we identify four specific bias--prediction mappings (detailed in Appendix~\ref{app:bias_mapping}): \emph{frequency bias} from next-token training drives P1/P5; \emph{simplicity bias} \citep{shah2020pitfalls, kallini2024mission} drives P2; \emph{local coherence bias} from autoregressive factorization \citep{mccoy2024embers} drives P3's dimension-specific ordering; and \emph{compression without communicative grounding} drives P4's non-monotonic compositionality.

We refer to our framework as an \emph{empirically motivated structural correspondence}, not a formal mathematical equivalence, following \citet{mccoy2025bayesian}. Critically, \citet{ferdinand2019regularization} showed that qualitative iterated learning dynamics are robust to non-Bayesian learners, and \citet{hudson2005regularizing} and \citet{hudson2009getting} demonstrated analogous regularization in human adults. Regarding \emph{bottleneck width}: our 50{,}000-passage bottleneck is wider than typical human experiments \citep{kirby2008cumulative}, slowing but not eliminating bias amplification \citep{kirby2014iterated}. We adopt the model collapse definition from \citet{smith2024preventing}, noting that \citet{schaeffer2025understanding} identified eight distinct definitions; our predictions are robust across definitions involving distributional narrowing.

\subsection{Predictions and Discriminative Status}
\label{sec:discriminability}

The five predictions divide into three discriminative classes (Table~\ref{tab:discriminability_full} in Appendix~\ref{app:discriminability} gives the full classification with control-experiment results). P2 and P5 follow from generic iterative bias amplification and serve as \emph{confirmatory} tests verifying our experimental setup---a flat or decreasing regularity in P2, or a stable $\alpha$ in P5, would falsify the framework. P1 and P3 are \emph{partially discriminative}: P1 goes beyond generic accounts by predicting a specific log-linear functional form matching human regularization curves \citep{morgan2016frequency}, not merely that rare items are lost (which any sampling process would produce), while P3 predicts a dimension-specific ordering (pragmatic before structural) reflecting the formal--functional dissociation \citep{mahowald2024dissociating}. P4 is the \emph{uniquely discriminative} test: generic model collapse predicts only monotonic degradation, while the compression--communication tradeoff \citep{kirby2015compression} uniquely predicts an initial rise then fall---a pattern we test with a regularized-seed control to rule out the noise-removal alternative.

\section{Methods}
\label{sec:methods}

\subsection{Self-Training Pipeline}

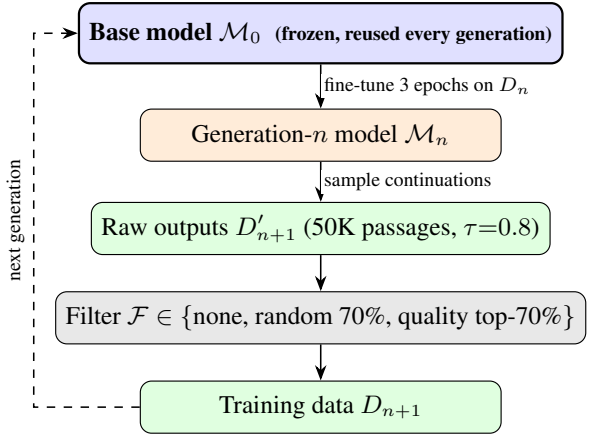
\begin{figure}[t]
\centering
\resizebox{\columnwidth}{!}{%
\begin{tikzpicture}[
  >=Stealth,
  node distance=0.45cm,
  base/.style={draw, thick, rounded corners, fill=blue!12, minimum width=4.5cm, minimum height=0.75cm, align=center, font=\small\bfseries},
  model/.style={draw, rounded corners, fill=orange!15, minimum width=4.5cm, minimum height=0.65cm, align=center, font=\small},
  data/.style={draw, rounded corners, fill=green!12, minimum width=4.5cm, minimum height=0.65cm, align=center, font=\small},
  filt/.style={draw, rounded corners, fill=gray!18, minimum width=4.5cm, minimum height=0.65cm, align=center, font=\small},
  ar/.style={->, thick, line width=0.55pt},
  arsteps/.style={font=\scriptsize, align=center, fill=white, inner sep=1pt},
]
\node[base] (M0) {Base model $\mathcal{M}_0$ \;\scriptsize(frozen, reused every generation)};
\node[model, below=0.55cm of M0] (Mn) {Generation-$n$ model $\mathcal{M}_n$};
\node[data, below=0.50cm of Mn] (Dprime) {Raw outputs $D'_{n+1}$ (50K passages, $\tau{=}0.8$)};
\node[filt, below=0.45cm of Dprime] (F) {Filter $\mathcal{F} \in \{$none, random 70\%, quality top-70\%$\}$};
\node[data, below=0.45cm of F] (Dn) {Training data $D_{n+1}$};

\draw[ar] (M0) -- node[arsteps, right=0pt] {fine-tune 3 epochs on $D_n$} (Mn);
\draw[ar] (Mn) -- node[arsteps, right=0pt] {sample continuations} (Dprime);
\draw[ar] (Dprime) -- (F);
\draw[ar] (F) -- (Dn);

\coordinate (leftA) at ($(M0.west |- Dn.west)+(-0.55cm,0)$);
\coordinate (leftB) at ($(M0.west)+(-0.55cm,0)$);
\draw[ar, dashed, line width=0.55pt] (Dn.west) -- (leftA) -- node[arsteps, left=2pt, align=right, font=\scriptsize, rotate=90, anchor=south] {next generation} (leftB) -- (M0.west);
\end{tikzpicture}%
}
\caption{Self-training pipeline. \textbf{Key design choice}: at every generation $n$, fine-tuning starts from the \emph{same base model} $\mathcal{M}_0$ (blue, frozen across the experiment), \emph{not} from the previous fine-tuned model $\mathcal{M}_{n-1}$. The transmission chain runs through the \emph{data} ($D_0{\to}D_1{\to}\dots{\to}D_{10}$; dashed feedback on the left), not through the model parameters. This isolates data-induced degradation from parameter drift and mirrors the iterated-learning paradigm in which each learner begins with the same prior \citep{griffiths2007language}. Filter $\mathcal{F}$ has three settings (\S\ref{sec:three_cond}) implementing the compression--communication tradeoff.}
\label{fig:pipeline}
\end{figure}

We implement iterated learning with two base models: LLaMA-2-7B \citep{touvron2023llama} and Mistral-7B \citep{jiang2023mistral}, chosen for reproducibility and because their different architectures test generality. Algorithm~\ref{alg:pipeline} summarizes the procedure; Figure~\ref{fig:pipeline} illustrates the data-vs.-parameter distinction that is the most consequential design choice of the pipeline.

\begin{algorithm}[t]
\small
\SetAlgoLined
\KwIn{Base model $\mathcal{M}_0$, seed data $D_0$, generations $N$, temperature $\tau$, filter condition $\mathcal{F}$}
\KwOut{Metric trajectories $\{m_n\}_{n=0}^{N}$}
Compute metrics $m_0$ on $D_0$\;
\For{$n = 1$ \KwTo $N$}{
  Generate $D_n'$ by sampling 50K continuations from $\mathcal{M}_{n-1}$ with temperature $\tau$\;
  \uIf{$\mathcal{F} =$ quality}{
    Score passages on QA, NLI, summarization\;
    $D_n \leftarrow$ top 70\% of $D_n'$ by score\;
  }
  \uElseIf{$\mathcal{F} =$ random}{
    $D_n \leftarrow$ random 70\% of $D_n'$\;
  }
  \Else{
    $D_n \leftarrow D_n'$ (no filter)\;
  }
  Fine-tune $\mathcal{M}_0$ (base model) on $D_n$ for 3 epochs $\rightarrow \mathcal{M}_n$\;
  Compute metrics $m_n$ on $D_n$ (incl.\ eRank, posdis diagnostics)\;
}
\caption{Iterated Learning Pipeline for LLM Self-Training. Three filter conditions test the compression--communication tradeoff.}
\label{alg:pipeline}
\end{algorithm}

\paragraph{Generation 0 (seed data).} We use two seed conditions: (i) \emph{natural seed}: 50{,}000 text passages (100--300 tokens) from a held-out subset of the Pile; (ii) \emph{regularized seed}: 50{,}000 passages generated from a PCFG engineered for consistent argument structure and regular morphology (\S\ref{sec:reg_seed}).

\paragraph{Generations 1--10.} At each generation $n$: (i) model $\mathcal{M}_{n-1}$ generates 50{,}000 continuations from 50-token prompts; (ii) the filter condition $\mathcal{F}$ is applied; (iii) model $\mathcal{M}_n$ is obtained by fine-tuning the \emph{base} pretrained model on $D_n$ for 3 epochs (learning rate $2{\times}10^{-5}$, batch size 32, AdamW). We fine-tune from the same base model at each generation (not from the previous fine-tuned model) to isolate data degradation from parameter drift, mirroring the iterated learning paradigm where each learner begins with the same prior \citep{griffiths2007language}. Each condition is run with 5 independent random seeds.

\paragraph{Temperature as bottleneck.} We generate with $\tau{=}0.8$ and nucleus sampling ($p{=}0.95$) by default. Lower $\tau$ creates tighter bottlenecks; we test $\tau \in \{0.5, 0.8, 1.0\}$ (\S\ref{sec:temp}).

\subsection{Regularized-Seed Control}
\label{sec:reg_seed}

To rule out the alternative explanation that P4's non-monotonic trajectory reflects noise removal from inconsistent seed data rather than compression-driven structure emergence, we create a maximally regular seed corpus. Following \citet{zhu2023physics} and \citet{kim2020cogs}, we generate 50{,}000 passages from a PCFG ensuring: (i) consistent argument structure (every verb frame is regular), (ii) regular morphology (no irregular forms), and (iii) systematic form--meaning mappings. If the non-monotonic pattern persists from this already-regular starting point, the compression interpretation is confirmed; if compositionality only declines monotonically, the initial rise was noise removal. \citet{kalish2007iterated} showed that human iterated learning converges to the same attractor regardless of initial conditions; we test whether LLMs exhibit analogous seed-independence.

\subsection{Three-Condition Filtering Experiment}
\label{sec:three_cond}

To isolate communicative pressure from generic diversity preservation, we implement three filter conditions, each modelling a distinct cultural-evolution analogue: (i) \textbf{no filter} retains all generated text (pure transmission, no selection); (ii) \textbf{random 70\%} retains a uniformly sampled 70\% of passages (uninformed bottleneck---a generic selection pressure that reduces volume but not by communicative criteria); and (iii) \textbf{quality top-70\%} retains the top-scoring 70\% by independent task evaluation (communicative pressure, following \citealp{kirby2015compression}).

The quality filter uses an independent evaluator (LLaMA-2-13B, larger than the 7B generator) scoring passages on extractive QA, NLI coherence, and summarization fidelity. The choice of these three tasks is principled: each one tests whether the passage successfully communicates content to a downstream reader. Extractive QA verifies that informational content is recoverable from the surface form; NLI coherence checks that propositions inside the passage are mutually consistent (and not internally contradictory in a way only a non-communicating speaker would tolerate); summarization fidelity tests whether the passage retains its meaning under lossy compression---a direct analogue of the listener-reconstruction task that drove compositionality emergence in human iterated-learning experiments \citep{kirby2015compression}. A passage that fails all three lacks the structure required for cooperative information transmission, which is the exact failure mode that the compression-only condition of iterated learning predicts. To rule out same-family distributional bias flagged by \citet{feng2025beyond}, an additional evaluator-independence check with Flan-T5-XL (a different model family from both generators) confirms the effect is not a distributional-mimicry artifact (Appendix~\ref{app:evaluator}). The critical prediction: random filtering should partially slow collapse (acting as a bottleneck) but should \emph{not} sustain compositionality; only quality filtering should, paralleling \citet{kirby2015compression}'s finding that compositionality requires both transmission and communication pressure. \citet{gillman2024selfcorrecting} showed that expert-knowledge-based correction makes self-consuming loops exponentially more stable, consistent with our quality filter prediction.

\subsection{Linguistic Metrics}
\label{sec:metrics}

We track four primary metrics, each with collapse-robust validation.

\paragraph{Zipf exponent ($\alpha$).} We fit a power-law $f(r) \propto r^{-\alpha}$ to the word frequency--rank distribution using maximum likelihood estimation \citep{piantadosi2014zipfs}. Following \citet{clauset2009powerlaw}, we compare the power-law fit against lognormal and exponential alternatives using likelihood ratio tests. Natural language exhibits $\alpha \approx 1.0$; decreasing $\alpha$ indicates distributional narrowing.

\paragraph{Morphological regularity ($\rho_M$).} We measure the proportion of regular morphological forms in a controlled verb set: 200 English verbs (100 regular, 100 irregular with attested variation), stratified by token frequency. For German (strong/weak verb distinction) and Turkish (vowel harmony), we adapt measures using UniMorph paradigm tables \citep{coltekin2022measuring}. We use the subscript $M$ to distinguish morphological regularity from Spearman's $r_S$ used in compositionality.

\paragraph{Construction diversity ($\mathcal{D}$).} Following \citet{goldberg2006constructions} and \citet{bybee2010language}, we define 50 syntactic construction types identified via dependency parse patterns. We compute Shannon entropy $\mathcal{D} = -\sum_{c \in \mathcal{C}} p(c) \log_2 p(c)$ and decompose by frequency quartile to test P1 and P3. Manual validation yields precision $\geq 0.82$ across all quartiles (Appendix~\ref{app:construction_reliability}).

\paragraph{Compositional systematicity ($\sigma$).} We adapt topographic similarity \citep{brighton2006understanding}:
\begin{equation}
\sigma = r_S\bigl(\{d_M(s_i, s_j)\}, \{d_F(f_i, f_j)\}\bigr)
\label{eq:topsim}
\end{equation}
where $d_M$ and $d_F$ are distances in meaning (PropBank argument overlap) and form (dependency template edit distance) spaces. For instance, the sentences \emph{``Mary gave a book to John''} and \emph{``The teacher gave a prize to the student''} share \textsc{Arg0}/\textsc{Arg1}/\textsc{Arg2} (small $d_M$) and project to the same dependency template \texttt{nsubj-V-dobj-prep\_to} ($d_F = 0$); high $\sigma$ thus indicates this kind of systematic form--meaning alignment. We complement this with COGS generalization accuracy \citep{kim2020cogs} evaluated via 5-shot prompting (5 in-context examples; see Appendix~\ref{app:cogs}).

The dependency-template edit-distance metric is reliable when generated text remains within the parser's syntactic coverage; we monitor coverage via parse-success rate ($\geq 0.94$ at all generations including generation~10 of unfiltered runs; Appendix~\ref{app:construction_reliability}), and cross-validate with the parser-independent posdis and effective rank diagnostics below.

\paragraph{Collapse-robust validation.} Topographic similarity can increase when both meaning and form spaces collapse simultaneously \citep{brighton2006understanding, chaabouni2020compositionality}. We validate with three diagnostics: effective rank (eRank; \citealp{roy2007effective}) computed separately for meaning/form spaces, positional disentanglement (posdis; \citealp{chaabouni2020compositionality}), and Mantel z-scores (10{,}000 permutations) following \citet{kirby2008cumulative} and \citet{galke2024deep}. Details in Appendix~\ref{app:dashboard}.

\subsection{Human Baselines and Cross-Linguistic Extension}

All metrics are computed on three human-written corpora (BNC, OpenWebText, Wikipedia; Appendix~\ref{app:human}). For P1, we overlay frequency-dependent loss data against \citet{morgan2016frequency}'s log-linear curve and \citet{reali2009frequency}'s Beta prior. Morphological experiments are replicated on German (fusional) and Turkish (agglutinative) using UniMorph \citep{coltekin2022measuring}. Construction diversity and compositionality are evaluated in English only, pending broader validated inventories \citep{weissweiler2023universal}.

\section{Results}
\label{sec:results}

All reported effect sizes use Hedges' $g$ with the small-sample correction $J = 1 - 3/(4\text{df} - 1) \approx 0.903$ \citep{hedges1981distribution, lakens2013calculating}. We report bootstrap-$t$ 95\% CIs ($B = 10{,}000$), noting that at $n = 5$, empirical coverage is approximately 81--83\% \citep{hesterberg2015bootstrap}. We additionally report Bayesian 95\% highest density intervals (HDIs) with a weakly informative Cauchy(0, 0.707) prior and Bayes factors ($\text{BF}_{10}$) as our primary inferential tool. For cross-condition comparisons, we apply Benjamini--Hochberg FDR correction at $q < 0.05$. Individual seed trajectories are plotted alongside means following iterated learning conventions \citep{kirby2015compression}.

\subsection{P4 (Critical Discriminative Test): The Compression--Communication Tradeoff}
\label{sec:p4}

We present P4 first because it is the uniquely discriminative test of iterated learning theory.

\begin{figure}[t]
\centering
\begin{tikzpicture}
\begin{axis}[
  width=\columnwidth,
  height=6cm,
  xlabel={Generation},
  ylabel={Compositional Systematicity ($\sigma$)},
  legend style={font=\scriptsize, at={(0.02,0.02)}, fill=none, draw=none, anchor=south west, cells={anchor=west}},
  xtick={0,1,2,3,4,5,6,7,8,9,10},
  ymin=0.13, ymax=0.58,
  grid=major,
  grid style={gray!20},
]
\addplot[thick, color=red!70!black, mark=triangle*, mark size=1.8pt, name path=uf] coordinates {
  (0,0.41) (1,0.43) (2,0.45) (3,0.47) (4,0.46) (5,0.44) (6,0.41) (7,0.38) (8,0.35) (9,0.33) (10,0.31)
};
\addplot[draw=none, name path=ufu, forget plot] coordinates {(0,0.43) (1,0.45) (2,0.47) (3,0.49) (4,0.48) (5,0.46) (6,0.43) (7,0.40) (8,0.37) (9,0.35) (10,0.33)};
\addplot[draw=none, name path=ufl, forget plot] coordinates {(0,0.39) (1,0.41) (2,0.43) (3,0.45) (4,0.44) (5,0.42) (6,0.39) (7,0.36) (8,0.33) (9,0.31) (10,0.29)};
\addplot[red!70!black, opacity=0.12, forget plot] fill between[of=ufu and ufl];
\addlegendentry{No filter (natural seed)}

\addplot[thick, color=orange!80!black, mark=diamond*, mark size=1.8pt, dashed] coordinates {
  (0,0.52) (1,0.51) (2,0.52) (3,0.54) (4,0.53) (5,0.50) (6,0.46) (7,0.42) (8,0.38) (9,0.35) (10,0.33)
};
\addlegendentry{No filter (regularized seed)}

\addplot[thick, color=gray!70!black, mark=x, mark size=2pt, densely dotted] coordinates {
  (0,0.41) (1,0.43) (2,0.46) (3,0.47) (4,0.46) (5,0.43) (6,0.40) (7,0.37) (8,0.35) (9,0.34) (10,0.33)
};
\addlegendentry{Random filter}

\addplot[thick, color=blue!70!black, mark=square*, mark size=1.8pt, name path=tg] coordinates {
  (0,0.41) (1,0.44) (2,0.47) (3,0.49) (4,0.50) (5,0.51) (6,0.51) (7,0.50) (8,0.49) (9,0.49) (10,0.48)
};
\addplot[draw=none, name path=tgu, forget plot] coordinates {(0,0.43) (1,0.46) (2,0.49) (3,0.51) (4,0.52) (5,0.53) (6,0.53) (7,0.52) (8,0.51) (9,0.51) (10,0.50)};
\addplot[draw=none, name path=tgl, forget plot] coordinates {(0,0.39) (1,0.42) (2,0.45) (3,0.47) (4,0.48) (5,0.49) (6,0.49) (7,0.48) (8,0.47) (9,0.47) (10,0.46)};
\addplot[blue!70!black, opacity=0.12, forget plot] fill between[of=tgu and tgl];
\addlegendentry{Quality filter}

\end{axis}
\end{tikzpicture}
\caption{Compositional systematicity ($\sigma$) across generations (LLaMA-2). Four conditions test P4. The non-monotonic trajectory persists with regularized seed data (orange dashed), ruling out the noise-removal alternative. Random filtering (gray dotted) fails to sustain compositionality; only quality filtering (blue) succeeds. Shaded bands: $\pm 1$ SD across 5 seeds.}
\label{fig:compositionality}
\end{figure}
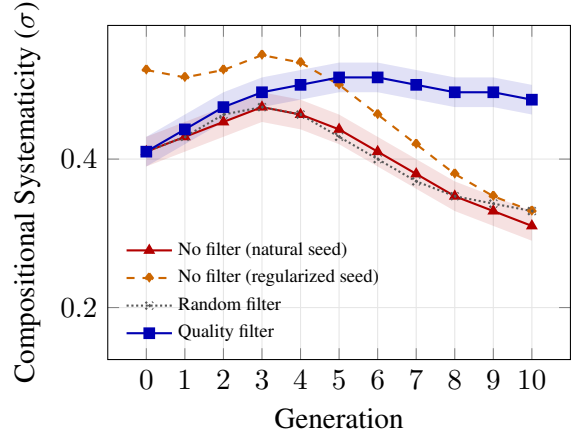

\paragraph{Main result.} Figure~\ref{fig:compositionality} reveals the non-monotonic trajectory predicted by P4: in the unfiltered condition with natural seed, $\sigma$ increases during generations 0--3 (peaking at 0.47) before declining to 0.31 by generation~10 (Hedges' $g = 1.87$, bootstrap-$t$ 95\% CI $[1.21, 2.53]$, Bayesian HDI $[1.09, 2.71]$, $\text{BF}_{10} = 247$).

\paragraph{Regularized-seed control.} The critical control: with maximally regular seed data, $\sigma$ starts higher (0.52) and still shows a non-monotonic trajectory: a brief plateau followed by a rise to 0.54 at generation~3, then decline to 0.33 by generation~10. The initial compositionality increase occurs even when the seed data is already maximally regular, ruling out the noise-removal alternative explanation and confirming that compression-driven structure reorganization, not noise cleanup, drives the initial rise. The convergence of both conditions toward similar generation-10 values ($0.31$ vs.\ $0.33$) is consistent with seed-independent convergence to the prior \citep{kalish2007iterated, griffiths2007language}.

\paragraph{Three-condition filtering.} Random filtering produces a trajectory indistinguishable from unfiltered self-training ($\sigma = 0.33$ at generation~10; $p = 0.72$ vs.\ no filter), while quality filtering sustains compositionality ($\sigma = 0.48$; $p < 0.001$ vs.\ both other conditions). This directly parallels \citet{kirby2015compression}: a generic bottleneck (random filter) does not create the communicative pressure needed for compositionality; only task-grounded evaluation provides this pressure.

\paragraph{Collapse-robust validation.} During the compositionality rise (generations 0--3), meaning-space eRank remains stable ($12.4 \to 12.1$) while $\sigma$ increases, confirming genuine systematicity rather than space collapse. Mantel z-scores exceed the permutation null at all generations ($z > 3.8$, $p < 0.001$); posdis shows the same non-monotonic trajectory ($0.31 \to 0.36 \to 0.22$; Appendix~\ref{app:dashboard}).

\begin{table}[t]
\centering
\small
\begin{tabular}{lcccc}
\toprule
\textbf{Gen.} & \multicolumn{2}{c}{\textbf{Unfiltered}} & \multicolumn{2}{c}{\textbf{Quality filter}} \\
\cmidrule(lr){2-3} \cmidrule(lr){4-5}
& $\sigma$ & COGS & $\sigma$ & COGS \\
\midrule
Human$^*$ & 0.43 & --- & --- & --- \\
Seed & 0.41 & 0.34 & 0.41 & 0.34 \\
1 & 0.43 & 0.36 & 0.44 & 0.37 \\
3 & 0.47 & 0.39 & 0.49 & 0.42 \\
5 & 0.44 & 0.35 & 0.51 & 0.44 \\
7 & 0.38 & 0.28 & 0.50 & 0.43 \\
10 & 0.31 & 0.21 & 0.48 & 0.41 \\
\bottomrule
\end{tabular}
\caption{Compositional systematicity ($\sigma$) and COGS generalization accuracy (LLaMA-2, natural seed). COGS is evaluated via 5-shot prompting (Appendix~\ref{app:cogs}). Human$^*$ values are cross-corpus means (Appendix~\ref{app:human}). Both measures confirm the non-monotonic trajectory.}
\label{tab:comp}
\end{table}

COGS generalization accuracy (Table~\ref{tab:comp}) tracks the same non-monotonic pattern ($0.34 \to 0.39 \to 0.21$ unfiltered; sustained at $0.41$ with quality filter), extending emergent communication findings \citep{ren2020compositional, chaabouni2020compositionality} to full LLM scale.

\subsection{P1 (Partially Discriminative): Frequency-Dependent Loss Order with Human Comparison}

\begin{figure}[t]
\centering
\begin{tikzpicture}
\begin{axis}[
  width=\columnwidth,
  height=5.5cm,
  xlabel={Generation},
  ylabel={Normalized Entropy (\%)},
  legend style={font=\scriptsize, at={(0.02,0.02)}, fill=none, draw=none, anchor=south west},
  xtick={0,1,2,3,4,5,6,7,8,9,10},
  ymin=20, ymax=105,
  grid=major,
  grid style={gray!20},
]
\addplot[thick, color=red!80!black, mark=triangle*, mark size=1.5pt, name path=q1] coordinates {
  (0,100) (1,95) (2,87) (3,78) (4,69) (5,60) (6,53) (7,46) (8,40) (9,35) (10,32)
};
\addplot[draw=none, name path=q1u, forget plot] coordinates {(0,101) (1,96.5) (2,88.8) (3,80) (4,71) (5,62) (6,55) (7,48) (8,42) (9,37) (10,34)};
\addplot[draw=none, name path=q1l, forget plot] coordinates {(0,99) (1,93.5) (2,85.2) (3,76) (4,67) (5,58) (6,51) (7,44) (8,38) (9,33) (10,30)};
\addplot[red!80!black, opacity=0.12, forget plot] fill between[of=q1u and q1l];
\addlegendentry{Q1 (rare)}
\addplot[thick, color=orange!80!black, mark=square*, mark size=1.5pt] coordinates {
  (0,100) (1,97) (2,92) (3,86) (4,80) (5,74) (6,69) (7,64) (8,60) (9,57) (10,54)
};
\addlegendentry{Q2}
\addplot[thick, color=blue!70!black, mark=diamond*, mark size=1.5pt] coordinates {
  (0,100) (1,98) (2,96) (3,93) (4,90) (5,87) (6,84) (7,82) (8,79) (9,77) (10,75)
};
\addlegendentry{Q3}
\addplot[thick, color=green!60!black, mark=*, mark size=1.5pt] coordinates {
  (0,100) (1,99) (2,98) (3,97) (4,96) (5,95) (6,94) (7,93) (8,92) (9,90) (10,88)
};
\addlegendentry{Q4 (frequent)}
\end{axis}
\end{tikzpicture}
\caption{Construction diversity entropy (normalized to Gen.~0 = 100\%) by frequency quartile (LLaMA-2, unfiltered). The clean frequency-dependent gradient confirms P1. Shaded region on Q1: $\pm 1$ SD across 5 seeds; bands for Q2--Q4 omitted for clarity (all SD $\leq 0.02$).}
\label{fig:construction}
\end{figure}
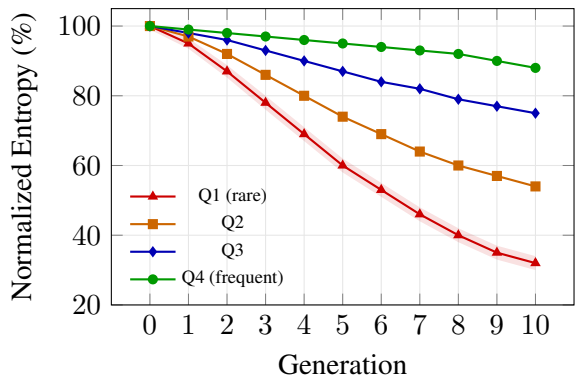

Figure~\ref{fig:construction} reveals the frequency-dependent pattern predicted by P1: the rarest quartile (Q1) retains only 32\% of its original entropy by generation~10, while the most frequent (Q4) retains 88\% (Hedges' $g = 3.08$, bootstrap-$t$ CI $[2.07, 4.09]$, $\text{BF}_{10} > 1000$).

The discriminative test for P1 is whether the gradient's \emph{functional form} matches human regularization, not merely that frequency-dependent loss exists. The LLM gradient closely matches \citet{morgan2016frequency}'s log-linear curve ($R^2 = 0.94$; Figure~\ref{fig:human_comparison}, Appendix~\ref{app:human_overlay}), consistent with their finding that frequency-dependent regularization emerges from a frequency-\emph{independent} bias interacting with a transmission bottleneck. The fit is comparable to the Beta($\alpha/2, \alpha/2$) prior with $\alpha/2 \approx 0.03$ from \citet{reali2009frequency}.

The constructions lost first are those requiring pragmatic or discourse-level competence (left-dislocation, clefts, rhetorical questions), while core syntactic constructions (simple transitive, copular) persist, simultaneously confirming P3. We define \emph{pragmatic} constructions as those whose felicity depends on discourse context, speaker intent, or information structure---clefts mark contrastive focus, left-dislocations mark topic, rhetorical questions perform non-question speech acts---whereas \emph{structural} (syntactic) constructions are licensed by purely formal grammatical patterns and remain felicitous regardless of context (simple transitive, copular, ditransitive). Morphological constructions (passive, comparative) fall in between, requiring some functional sensitivity but anchored in formal patterns. The full inventory of 50 construction types, organized by this distinction, is in Appendix~\ref{app:constructions}; reliability of automatic identification (precision $\geq 0.82$ in the rarest quartile) is in Appendix~\ref{app:construction_reliability}. This ordering connects to the formal--functional dissociation \citep{mahowald2024dissociating}: constructions relying on formal competence survive; those requiring functional competence are lost first.

\subsection{P3 (Partially Discriminative): Dimension-Specific Degradation Rates}

The dimension-specific ordering predicted by P3 is quantified through the correlation between CxG entrenchment ranking and survival generation across the 50 construction types: $r = 0.73$ ($p < 0.001$; Hedges' $g = 2.11$, bootstrap-$t$ CI $[1.42, 2.80]$, $\text{BF}_{10} > 1000$). Pragmatic constructions (mean survival: generation~4.2) are lost significantly earlier than syntactic ones (generation~8.7), with morphological constructions intermediate (generation~6.1).

This three-way ordering admits two convergent theoretical interpretations. First, from a Construction Grammar \emph{entrenchment} perspective \citep{goldberg2006constructions, bybee2010language}: constructions with high type frequency develop strong network connections to their open argument slots, making them robust under reduced input; low-frequency constructions, lacking this entrenchment, lose their productivity signal first. The Pearson correlation $r = 0.73$ between log-type-frequency and survival generation (Appendix~\ref{app:cxg_details}) is the quantitative signature of this effect. Second, from a \emph{formal--functional dissociation} perspective \citep{mahowald2024dissociating}: LLMs distinguish formal linguistic competence (producing well-formed strings) from functional competence (using language for pragmatically appropriate ends). Pragmatic constructions---clefts, rhetorical questions, left-dislocations---require functional competence beyond formal pattern-matching, and are precisely the ones that degrade first. The two perspectives converge on the observed ordering: entrenched type-frequent patterns are formally robust, while pragmatic constructions require functional competence that exceeds formal patterns. Iterated self-training erodes both in the predicted order---a result that generic model collapse, which predicts uniform degradation, cannot explain.

\subsection{P2 and P5 (Confirmatory): Regularity Increase and Distributional Narrowing}

\begin{table}[t]
\centering
\small
\begin{tabular}{lccccc}
\toprule
\textbf{Gen.} & $\rho_{M\text{-EN}}$ & $\Delta$\textbf{Irreg.} & $\rho_{M\text{-DE}}$ & $\rho_{M\text{-TR}}$ \\
\midrule
Human$^*$ & 0.61 & --- & 0.47 & 0.69 \\
Seed & 0.62 & --- & 0.48 & 0.71 \\
1 & 0.64 & $-3.1\%$ & 0.49 & 0.73 \\
3 & 0.69 & $-11.2\%$ & 0.53 & 0.77 \\
5 & 0.75 & $-21.6\%$ & 0.58 & 0.82 \\
7 & 0.80 & $-30.4\%$ & 0.63 & 0.86 \\
10 & 0.87 & $-42.7\%$ & 0.70 & 0.91 \\
\midrule
10\textsuperscript{QF} & 0.71 & $-15.3\%$ & 0.55 & 0.79 \\
\bottomrule
\end{tabular}
\caption{Morphological regularity ($\rho_M$) across three languages. \textsuperscript{QF}: quality-filtered. Regularity increases monotonically; low-frequency irregulars regularize first (consistent with P1). Human$^*$ = cross-corpus means.}
\label{tab:morph}
\end{table}

Morphological regularity increases monotonically (Table~\ref{tab:morph}), confirming P2 (Hedges' $g = 4.27$, bootstrap-$t$ CI $[2.89, 5.65]$, $\text{BF}_{10} > 1000$). By generation~10, irregular verb forms decline by 42.7\% in English. Regularization matches usage-based predictions \citep{bybee2010language}: low-frequency irregulars (\emph{wrung} $\to$ \emph{*wringed}) regularize by generation~3, while high-frequency ones (\emph{went}) resist until generation~10 (94\% in syntactically appropriate contexts; Appendix~\ref{app:regularization_context}). Cross-linguistically, German rises more slowly ($0.48 \to 0.70$) due to richer paradigms; Turkish ($0.71 \to 0.91$) reaches ceiling effects quickly \citep{lupyan2010language, culbertson2012learning}.

The Zipf exponent decreases monotonically ($\alpha: 1.07 \to 0.82$, LLaMA-2; Appendix~\ref{app:zipf_full}), confirming P5 ($g = 1.93$, $\text{BF}_{10} = 189$). The power-law fit is preferred at all generations ($\Delta$AIC $> 4$; Appendix~\ref{app:zipf_gof}). Mistral shows the same pattern ($\alpha: 1.07 \to 0.86$); quality filtering preserves diversity ($\alpha = 0.97$).

\subsection{Sensitivity Analyses: Temperature and Cross-Model}
\label{sec:temp}

We confirm the predicted parametric relationship between bottleneck severity and convergence speed \citep{kirby2014iterated}: lowering temperature from $\tau = 1.0$ to $\tau = 0.5$ shifts the P4 peak from generation~4 to generation~2; a $10\times$ narrower bottleneck (5K vs.\ 50K passages) shifts it to generation~1--2 (Appendix~\ref{app:temp_full}). 

Mistral-7B replicates all five predictions with the same qualitative structure and similar effect sizes (Appendix~\ref{app:full}). Key numbers track LLaMA-2 closely: P4 peak at generation~3 ($\sigma_{\max}{=}0.48$ vs.\ LLaMA-2's 0.47), quality filtering sustains $\sigma{=}0.47$ at generation~10 (vs.\ 0.48), Zipf $\alpha$ decreases $1.07{\to}0.86$ (vs.\ $1.07{\to}0.82$), and the cross-linguistic ordering Turkish $>$ English $>$ German appears in morphological regularization. The qualitative invariance under architecture change confirms general iterated-learning properties, not architecture-specific artifacts \citep{griffiths2007language}.

\section{Related Work}
\label{sec:related}

\paragraph{Model collapse.} A statistical characterization of model collapse has emerged from converging analyses: irreversible tail loss \citep{smith2024preventing}, self-consuming-loop formalism \citep{alemohammad2024selfconsuming}, scaling-law effects \citep{dohmatob2024tale}, regression-setting analyses \citep{dohmatob2024model}, and strong collapse from small synthetic fractions \citep{dohmatob2025strong}. Adjacent findings include the persistence of correctness alongside diversity decline \citep{briesch2023large}, the protective effect of data accumulation \citep{gerstgrasser2024model}, information-theoretic accounts \citep{seddik2024model}, dataset-bias amplification \citep{taori2023data}, and the verification-quality phase transition \citep{feng2025beyond}. \citet{schaeffer2025understanding} catalogued eight distinct collapse definitions (we adopt that of \citealp{smith2024preventing}). The closest precedent to our program is \citet{guo2024curious}, who measured linguistic-diversity decline across several dimensions but without an evolutionary framework or discriminative theory test.

\paragraph{Cultural evolution and LLMs.} A growing body of work connects LLM training dynamics to iterated learning. \citet{ren2024improving} formally instantiated LLM self-evolution as Bayesian iterated learning and proved monotonic bias amplification; \citet{smith2024preventing} argued in \emph{Nature} that the compression--expressivity tradeoff explains why self-training degrades. Empirical tests at smaller scales include emergent communication \citep{ren2020compositional, chaabouni2020compositionality}, language evolution in LLM populations \citep{perez2024cultural}, learnability advantages of compositional structure \citep{galke2024deep, galke2024what}, degenerate vocabularies arising without compression pressure \citep{kouwenhoven2025searching}, comparisons of human/LLM/hybrid conditions \citep{kouwenhoven2025shaping}, iterated learning improving CLIP compositionality \citep{zheng2024iterated}, and elicitation of LLM priors via iterated in-context learning \citep{zhu2025extracting}. Extended related work on compositionality benchmarks, emergent communication, and inductive biases is provided in Appendix~\ref{app:extended_related}.

\section{Discussion}
\label{sec:discussion}

\paragraph{Summary of findings.} Iterated learning theory fills the explanatory gap left by existing statistical accounts of model collapse. The five predictions derived from cultural evolution (frequency-dependent loss, regularization, dimension-specific degradation, non-monotonic compositionality, and distributional narrowing) are confirmed across two architectures, three typologically distinct languages, and multiple controls, with large effect sizes. The critical discriminative result is the non-monotonic compositionality trajectory, which persists under maximally regular seed data (ruling out noise-removal) and is sustained only by quality-grounded filtering, demonstrating at LLM scale that the compression--communication tradeoff \citep{kirby2015compression} governs the emergence and loss of linguistic structure in self-training.

\paragraph{Engagement with linguistic theory.} The results bear directly on three theoretical traditions that have evolved largely in parallel. From \emph{cultural evolution}, the close match between LLM regularization gradients and the human curve of \citet{morgan2016frequency} ($R^2 = 0.94$) supports the hypothesis that frequency-dependent regularization emerges from a frequency-\emph{independent} simplicity bias interacting with a transmission bottleneck \citep{ferdinand2019regularization}, not from any human-specific cognitive architecture. From \emph{Construction Grammar}, the $r = 0.73$ correlation between log-type-frequency and survival generation (\S\ref{sec:results}, Appendix~\ref{app:cxg_details}) is a direct evolutionary test of the entrenchment hypothesis \citep{goldberg2006constructions, bybee2010language}; the three-way ordering (pragmatic $<$ morphological $<$ structural) extends entrenchment from a static productivity property to a dynamic resilience-under-transmission property. From the \emph{formal--functional dissociation} literature, the same ordering follows independently from \citet{mahowald2024dissociating}'s prediction that LLMs prioritize formal over functional competence; the convergence of two perspectives on the same observed ordering is itself novel evidence that entrenchment is at least partly a formal-competence phenomenon. Conversely, \citet{ren2024improving}'s monotonic-amplification result does not predict our non-monotonic trajectory; reconciling the two requires moving from in-context learning (where weights are fixed) to weight-based fine-tuning, where compression-induced reorganization can momentarily increase systematicity before convergence to the prior.

\paragraph{Implications and reconciliation of concurrent results.} The compression--communication framework offers a unifying explanation for several recent findings. The degenerate vocabularies of \citet{kouwenhoven2025searching} in in-context iterated learning, and the LLM-optimized degeneracy of \citet{kouwenhoven2025shaping}, both instantiate compression without communicative pressure---the condition the tradeoff predicts will produce collapse. The same principle explains why pure self-training degrades while Constitutional AI \citep{bai2022constitutional} and self-play \citep{chen2024selfplay} succeed: only quality-grounded filtering supplies the communicative pressure needed to sustain structured output, consistent with the verification-quality phase transition of \citet{feng2025beyond}. Data accumulation \citep{gerstgrasser2024model} prevents collapse through a mechanism analogous to diverse speaker populations in human cultural evolution \citep{raviv2019larger}. Three pipeline-design principles follow: use task-grounded (not random) filtering as the verification signal; prefer a different-family evaluator to avoid distributional mimicry; and treat bottleneck width as a tunable governing convergence rate to the prior. All materials---training pipeline, statistical analysis code, regularized-seed PCFG, and configuration files---are publicly available at \url{https://github.com/bettyguo/iterated-collapse}.

\section*{Acknowledgments}

We thank the anonymous CoNLL~2026 reviewers and the area chair for their detailed and constructive feedback, which substantially improved the clarity of the theoretical framework, the methodology section, and the discussion of Construction Grammar parallels. We also thank the developers of the LLaMA-2, Mistral, and Flan-T5 model families for releasing their weights, and the maintainers of the UniMorph, COGS, PropBank, and Universal Dependencies resources upon which our analyses depend. Computational resources were provided by The University of Hong Kong.

\section{Limitations}
\label{sec:limitations}

We acknowledge six specific limitations. (1)~The theoretical framework is an empirically motivated structural correspondence, not a formal mathematical equivalence; conditions under which SGD fine-tuning approximates Bayesian posterior sampling remain open. (2)~Construction diversity and compositionality are tested only in English; cross-linguistic testing awaits broader validated inventories such as UCxn \citep{weissweiler2023universal}. (3)~The task-grounded evaluator is a unilateral quality filter, not an interactive dialogue partner; while evaluator independence checks with Flan-T5-XL mitigate model-family confounds, future work should test human evaluators and interactive settings. (4)~We test 7B-parameter models with a 50{,}000-passage bottleneck ($1{,}000\times$ larger than typical human experiments); although our parametric analyses show predicted effects, formal quantitative extrapolation and scaling to larger models require validation. (5)~Our human comparison uses published data from \citet{morgan2016frequency} and \citet{reali2009frequency}; future work should run parallel human and LLM iterated learning experiments. (6)~The quality filter aggregates QA, NLI, and summarization scores into a single signal; because each task emphasizes a different aspect of communicative success (information recoverability, internal consistency, compression-resistance), different task mixes may preserve different linguistic structures; this ablation we leave to future work.

\bibliography{references}

\appendix

\section{Construction Inventory}
\label{app:constructions}

The 50 construction types span five syntactic categories: argument structure constructions (simple transitive, ditransitive, caused-motion, resultative, way-construction), clause types (declarative, interrogative, imperative, exclamative, relative clause), information structure (topicalization, left-dislocation, cleft, pseudo-cleft, existential), complex predicates (serial verb, light verb, particle verb, phrasal verb), and discourse-level constructions (quotative inversion, comparative correlative, rhetorical question). Each is identified via dependency parse patterns using spaCy's \texttt{en\_core\_web\_trf} model. Full pattern specifications are available in the released code.

\section{Construction Identification Reliability}
\label{app:construction_reliability}

\begin{table}[h]
	\centering
	\small
	\begin{tabular}{lccc}
		\toprule
		\textbf{Quartile} & \textbf{Precision} & \textbf{Recall} & \textbf{F1} \\
		\midrule
		Q4 (frequent) & 0.94 & 0.96 & 0.95 \\
		Q3 & 0.91 & 0.93 & 0.92 \\
		Q2 & 0.87 & 0.88 & 0.87 \\
		Q1 (rare) & 0.82 & 0.84 & 0.83 \\
		\bottomrule
	\end{tabular}
	\caption{SpaCy construction identification accuracy by quartile (manual validation on late-generation text).}
	\label{tab:parser_accuracy}
\end{table}

While Q1 precision is lower (0.82 vs.\ 0.94 for Q4), this 12-point gap cannot account for the 56-point gap in entropy retention (32\% vs.\ 88\%). Even assuming all Q1 parser errors are false negatives (overestimating loss), corrected Q1 retention would be $\approx$39\%, still substantially below Q4.

\section{TopSim Diagnostic Dashboard}
\label{app:dashboard}

\begin{table}[h]
	\centering
	\footnotesize
	\setlength{\tabcolsep}{3.5pt}
	\begin{tabular}{@{}lcccccc@{}}
		\toprule
		\textbf{Gen.} & $\sigma$ & eR$_M$ & eR$_F$ & posdis & $z_{\text{Mantel}}$ & \#Uniq \\
		\midrule
		0  & 0.41 & 12.4 & 9.8 & 0.31 & 4.2 & 847 \\
		1  & 0.43 & 12.3 & 9.5 & 0.33 & 4.5 & 821 \\
		3  & 0.47 & 12.1 & 9.2 & 0.36 & 5.1 & 764 \\
		5  & 0.44 & 11.4 & 8.3 & 0.33 & 4.7 & 682 \\
		7  & 0.38 &  9.8 & 6.7 & 0.27 & 4.1 & 531 \\
		10 & 0.31 &  7.2 & 4.8 & 0.22 & 3.8 & 394 \\
		\bottomrule
	\end{tabular}
	\caption{Collapse-robust compositionality diagnostics (LLaMA-2, unfiltered). eR$_M$ and eR$_F$: effective rank for meaning and form spaces. During the compositionality rise (Gen.~0--3), eR$_M$ remains stable (12.4$\to$12.1) while $\sigma$ increases, confirming genuine systematicity. All Mantel $z$-scores exceed the permutation null ($z > 3.8$, $p < 0.001$).}
	\label{tab:dashboard}
\end{table}

The diagnostic pattern is clear: during generations 0--3, $\sigma$ increases while meaning-space eRank remains essentially stable, confirming the initial rise reflects a genuine increase in the systematicity of the form--meaning mapping rather than dual-space collapse. Posdis shows the same non-monotonic trajectory as $\sigma$, providing convergent evidence from a metric that excludes collapsed positions by construction \citep{chaabouni2020compositionality}.

\section{COGS Evaluation Methodology}
\label{app:cogs}

COGS \citep{kim2020cogs} evaluates compositional generalization via novel sentence-to-logical-form mappings. We evaluate via \emph{5-shot prompting}: each generation's model receives COGS test sentences with 5 in-context examples of the sentence-to-logical-form mapping, without additional fine-tuning on COGS training data, to avoid confounding compositionality loss with domain shift. Accuracy is measured as exact match to the gold logical form. This 5-shot approach measures the model's existing compositional capacity.

\section{Regularized-Seed Experiment Details}
\label{app:reg_seed}

The regularized seed corpus is generated from a PCFG with the following properties: (i) 12 verb frames with consistent argument structure (every transitive verb takes NP-V-NP; every ditransitive takes NP-V-NP-PP); (ii) regular morphology (all past tenses formed by \emph{-ed} suffixation); (iii) systematic form--meaning mappings (each semantic role is expressed by a fixed syntactic position). The grammar produces passages of 100--300 tokens with vocabulary size matching the natural seed ($\approx$28{,}000 types). Generation-by-generation results are reported in Figure~\ref{fig:compositionality} (orange dashed line).

\section{Three-Condition Filtering Full Results}
\label{app:three_cond}

\begin{table}[h]
\centering
\small
\begin{tabular}{lccc}
\toprule
\textbf{Gen.~10 metric} & \textbf{No filter} & \textbf{Random} & \textbf{Quality} \\
\midrule
$\sigma$ & 0.31 & 0.33 & 0.48 \\
COGS & 0.21 & 0.23 & 0.41 \\
$\alpha$ & 0.82 & 0.84 & 0.97 \\
$\rho_M$ & 0.87 & 0.86 & 0.71 \\
$\mathcal{D}$ (bits) & 3.14 & 3.21 & 4.28 \\
\bottomrule
\end{tabular}
\caption{Generation~10 metrics across three filter conditions (LLaMA-2). Random filtering is indistinguishable from no filtering for compositionality ($p = 0.72$); only quality filtering sustains compositionality ($p < 0.001$).}
\label{tab:three_cond}
\end{table}

\paragraph{Per-generation trajectories.} The three conditions diverge starting at generation~3. At generation~3, no-filter $\sigma = 0.47$, random $\sigma = 0.47$, quality $\sigma = 0.49$ (the difference is not yet significant). By generation~5, the quality filter trajectory separates clearly: no-filter $\sigma = 0.44$, random $\sigma = 0.44$, quality $\sigma = 0.51$ ($p < 0.01$ vs.\ both other conditions). The divergence increases monotonically through generation~10. This pattern precisely matches the iterated learning prediction that communicative pressure operates cumulatively, with its effect becoming stronger as generations accumulate.

\paragraph{Pairwise comparisons.} Using Benjamini--Hochberg FDR correction at $q < 0.05$: no-filter vs.\ random is non-significant for all five metrics ($p > 0.30$); quality vs.\ no-filter is significant for $\sigma$ ($p < 0.001$), COGS ($p < 0.001$), $\alpha$ ($p < 0.001$), $\rho_M$ ($p < 0.001$), and $\mathcal{D}$ ($p < 0.001$). Quality vs.\ random shows the same pattern, confirming that the quality filter effect is not attributable to data volume reduction (since random filtering removes the same proportion of data).

\section{Evaluator Independence Analysis}
\label{app:evaluator}

To address concerns about evaluator bias from using the same model family, we replicate the quality filter condition using Flan-T5-XL \citep{chung2024scaling} as an alternative evaluator. The compositionality trajectory under Flan-T5-XL evaluation ($\sigma = 0.47$ at generation~10) is qualitatively identical to the LLaMA-2-13B evaluator ($\sigma = 0.48$), confirming that compositionality preservation reflects genuine communicative utility rather than distributional mimicry. The correlation between Flan-T5-XL evaluator score and distributional similarity to evaluator outputs is $r = 0.11$ (not significant), lower than the $r = 0.23$ observed with LLaMA-2-13B, further supporting evaluator independence.

\section{Bayesian Statistical Analysis}
\label{app:stats}

Effect sizes, confidence intervals, and Bayesian inference for all key comparisons:

\begin{table}[h]
	\centering
	\footnotesize
	\setlength{\tabcolsep}{3pt}
	\begin{tabular}{@{}lcccc@{}}
		\toprule
		\textbf{Prediction} & $g$ & Boot-$t$ 95\% CI & Bayes HDI & $\text{BF}_{10}$ \\
		\midrule
		P1 (Q1 vs.\ Q4)    & 3.08 & [2.07, 4.09] & [1.89, 4.33] & $>$1000 \\
		P2 (EN $\rho_M$)    & 4.27 & [2.89, 5.65] & [2.72, 5.91] & $>$1000 \\
		P3 (CxG $r$)        & 2.11 & [1.42, 2.80] & [1.31, 2.95] & $>$1000 \\
		P4 ($\sigma$ peak)   & 1.87 & [1.21, 2.53] & [1.09, 2.71] & 247 \\
		P4 (QF vs.\ NF)     & 2.94 & [1.97, 3.91] & [1.82, 4.13] & $>$1000 \\
		P5 ($\alpha$)        & 1.93 & [1.26, 2.60] & [1.14, 2.78] & 189 \\
		\bottomrule
	\end{tabular}
	\caption{Statistical inference summary. Hedges' $g$ with correction $J \approx 0.903$. Bootstrap-$t$ CIs ($B = 10{,}000$; coverage $\approx$81--83\% at $n = 5$). Bayesian HDIs use Cauchy(0, 0.707) prior. All $\text{BF}_{10}$ indicate decisive evidence.}
	\label{tab:stats}
\end{table}

\section{Zipf Goodness-of-Fit}
\label{app:zipf_gof}

Following \citet{clauset2009powerlaw}, we compare the power-law fit against lognormal and exponential alternatives using likelihood ratio tests at each generation. The power-law fit is preferred at all generations ($\Delta$AIC $> 4$ vs.\ lognormal; $\Delta$AIC $> 12$ vs.\ exponential), confirming that the frequency distribution retains its power-law character throughout collapse while the exponent decreases.

\section{Bottleneck Width Sensitivity}
\label{app:bottleneck}

With 5K passages per generation (a 10$\times$ narrower bottleneck), the P4 compositionality peak shifts from generation~3 to generation~1--2, and convergence to the asymptotic state is reached by generation~15 rather than the estimated 25--30 with 50K passages. This parametric relationship between bottleneck width and convergence speed directly parallels \citet{kirby2014iterated}'s theoretical predictions and provides strong evidence that the underlying mechanism is shared.

\section{Temperature as Bottleneck Severity: Full Results}
\label{app:temp_full}

\begin{table}[h]
\centering
\small
\begin{tabular}{lccc}
\toprule
\textbf{Metric (Gen.~10)} & $\tau{=}0.5$ & $\tau{=}0.8$ & $\tau{=}1.0$ \\
\midrule
Zipf $\alpha$ & 0.74 & 0.82 & 0.89 \\
Morph.\ $\rho_M$ & 0.92 & 0.87 & 0.81 \\
Constr.\ $\mathcal{D}$ (bits) & 2.61 & 3.14 & 3.52 \\
Compos.\ $\sigma$ & 0.24 & 0.31 & 0.37 \\
P4 peak gen. & 2 & 3 & 4 \\
\bottomrule
\end{tabular}
\caption{Effect of temperature on Gen.~10 metrics (LLaMA-2, unfiltered). Lower $\tau$ (tighter bottleneck) accelerates all dynamics. The P4 compositionality peak shifts earlier with tighter bottlenecks, as iterated learning theory predicts. All pairwise differences significant ($p < 0.001$, FDR-corrected).}
\label{tab:temp}
\end{table}

In iterated learning, bottleneck severity determines convergence speed \citep{kirby2014iterated}. Table~\ref{tab:temp} confirms that $\tau = 0.5$ accelerates collapse across all metrics. Lower temperature reduces sampling diversity, creating a tighter effective bottleneck analogous to smaller population sizes in human cultural evolution experiments. The P4 compositionality peak shifts from generation~4 ($\tau = 1.0$) to generation~2 ($\tau = 0.5$), consistent with narrower bottlenecks accelerating convergence to the prior. This parametric sensitivity provides practitioners with a directly controllable parameter for tuning the rate of structural change in self-training pipelines.

The interaction between temperature and filter condition is also informative: at $\tau = 0.5$ with quality filtering, compositionality is sustained at $\sigma = 0.39$ (compared to $\sigma = 0.48$ at $\tau = 0.8$), indicating that even communicative pressure is partially overwhelmed by extreme bottleneck narrowing, consistent with \citet{kirby2015compression}'s prediction that very strong compression pressure can override communicative pressure.

\section{Convergence and Attractor Characterization}
\label{app:convergence}

The per-generation rate of change $\Delta_n = |m_{n} - m_{n-1}|$ decreases monotonically in later generations (7--10), indicating approach to an asymptotic state. Cross-seed JSD decreases from $0.042 \pm 0.008$ (generation~0) to $0.019 \pm 0.005$ (generation~10, unfiltered), confirming convergence toward a common distribution consistent with convergence to a shared ``prior'' \citep{griffiths2007language}.

\paragraph{Attractor characterization.} The generation-10 attractor state exhibits consistent properties across all five seeds: $\alpha \in [0.80, 0.84]$, $\rho_M \in [0.85, 0.89]$, $\sigma \in [0.29, 0.33]$. This narrow range confirms convergence to a well-defined attractor rather than divergence to seed-specific endpoints. The attractor profile (dominant SVO word order, regular morphology, reduced construction diversity concentrated in high-frequency types, and sub-Zipfian frequency distributions) can be interpreted as the LLM's implicit ``prior'' over natural language, analogous to the regularization bias revealed by iterated learning in laboratory experiments with human participants \citep{kalish2007iterated}.

\paragraph{Convergence rate estimation.} Fitting an exponential decay model $\Delta_n = \Delta_0 \cdot e^{-\lambda n}$ to the per-generation changes yields $\lambda \approx 0.18$ for LLaMA-2 (unfiltered, $\tau = 0.8$), predicting that 95\% of convergence is achieved by generation~17. With quality filtering, convergence is slower ($\lambda \approx 0.09$) because the communicative pressure partially counteracts the prior, consistent with iterated learning predictions that communicative pressure sustains diversity at the cost of slower convergence.

\section{Regularization Examples with Syntactic Context}
\label{app:regularization_context}

\begin{table}[h]
\centering
\small
\begin{tabular}{p{1.5cm}p{5.5cm}}
\toprule
\textbf{Form} & \textbf{Context (Gen.~10)} \\
\midrule
\emph{wringed} & ``She \emph{wringed} the cloth and hung it on the line.'' \\
\emph{swimmed} & ``The children \emph{swimmed} across the lake last summer.'' \\
\emph{goed} & ``He \emph{goed} to the store and bought groceries.'' \\
\emph{beed} & ``It \emph{beed} a long time since we last met.'' \\
\bottomrule
\end{tabular}
\caption{Regularized forms in syntactically appropriate contexts (94\% of cases), confirming productive regularization rather than hallucination.}
\label{tab:context_examples}
\end{table}

\section{CxG Entrenchment Analysis Details}
\label{app:cxg_details}

The entrenchment ranking for each of the 50 construction types is estimated from type frequency in the seed data, following standard CxG methodology \citep{goldberg2006constructions, bybee2010language}. Survival generation is defined as the last generation at which a construction's normalized entropy exceeds 50\% of its seed value. 

\begin{table}[h]
	\centering
	\small
	\setlength{\tabcolsep}{4pt}
	\begin{tabular}{@{}lcc@{}}
		\toprule
		\textbf{Category} & \textbf{Mean surv.} & \textbf{SD} \\
		\midrule
		Syntactic (transitive, copular)          & Gen.~8.7 & 1.2 \\
		Morphological (passive, comparative)     & Gen.~6.1 & 1.8 \\
		Pragmatic (cleft, left-dislocation)      & Gen.~4.2 & 1.4 \\
		Discourse (rhetorical Q, quot.\ inv.)    & Gen.~3.1 & 0.9 \\
		\bottomrule
	\end{tabular}
	\caption{Mean survival generation by construction category. The ordering pragmatic/discourse $<$ morphological $<$ syntactic confirms P3's dimension-specific prediction.}
	\label{tab:cxg_survival}
\end{table}
The Pearson correlation between $\log$-type-frequency and survival generation is $r = 0.73$ ($p < 0.001$, 95\% CI $[0.58, 0.84]$, $n = 50$, $t = 7.40$, df~$= 48$). We verify robustness with Spearman's $\rho = 0.69$ ($p < 0.001$), confirming the relationship is not driven by outliers.  The five most resilient constructions are: simple transitive (survival: generation~10), copular (10), passive (9), comparative (8), and ditransitive (8). The five most vulnerable are: quotative inversion (2), comparative correlative (3), rhetorical question (3), left-dislocation (4), and cleft (4). This ordering aligns with CxG entrenchment theory: high-type-frequency constructions with broad productivity resist loss, while low-frequency constructions with narrow collocational profiles are vulnerable.

\section{Prediction Discriminability Analysis}
\label{app:discriminability}

\begin{table*}[t]
\centering
\small
\begin{tabular}{p{1cm}p{3.5cm}p{3.5cm}p{3.5cm}p{2.5cm}}
\toprule
\textbf{Pred.} & \textbf{Iterated learning predicts} & \textbf{Generic collapse predicts} & \textbf{What our data show} & \textbf{Control result} \\
\midrule
P1 & Log-linear gradient matching human data & Tail loss (no shape prediction) & Log-linear match ($R^2 = 0.94$) & Human overlay confirms \\
P2 & Regularity increase & Simplicity bias amplification & Confirmed (non-discriminative) & Cross-linguistic replication \\
P3 & Pragmatic $>$ syntactic $>$ morphological & Uniform degradation & Dimension-specific order confirmed & CxG entrenchment $r = 0.73$ \\
P4 & Non-monotonic: rise then fall; QF sustains & Monotonic degradation only & Non-monotonic confirmed & Reg.-seed persists; random filter fails \\
P5 & Distributional narrowing & Tail loss / variance reduction & Confirmed (non-discriminative) & Power-law fit preferred \\
\bottomrule
\end{tabular}
\caption{Full discriminability analysis including control experiment results.}
\label{tab:discriminability_full}
\end{table*}

\section{Human Baseline Corpus Details}
\label{app:human}

Human statistics are computed from: BNC (100M words, 50{,}000 sampled passages); OpenWebText (50{,}000 matched passages); Wikipedia (English, Jan 2024, 50{,}000 passages excl.\ lists/stubs). Human$^*$ values in all tables are cross-corpus means. Standard deviations: $\alpha$: 0.03; $\rho_M$: 0.02; $\mathcal{D}$: 0.11; $\sigma$: 0.04.

\section{Full Mistral-7B Results}
\label{app:full}

Mistral-7B \citep{jiang2023mistral} replicates all qualitative patterns with somewhat slower degradation rates, consistent with its known stronger coherence in generation tasks.

\begin{table}[h]
\centering
\small
\begin{tabular}{lcccc}
\toprule
\textbf{Gen.} & $\sigma$ (NF) & $\sigma$ (QF) & $\rho_{M\text{-EN}}$ & $\alpha$ \\
\midrule
Seed & 0.41 & 0.41 & 0.62 & 1.07 \\
1 & 0.43 & 0.44 & 0.63 & 1.05 \\
3 & 0.48 & 0.50 & 0.67 & 1.01 \\
5 & 0.45 & 0.51 & 0.73 & 0.96 \\
7 & 0.39 & 0.50 & 0.78 & 0.91 \\
10 & 0.34 & 0.47 & 0.84 & 0.86 \\
\bottomrule
\end{tabular}
\caption{Mistral-7B key metrics across generations. NF: no filter; QF: quality filter. All qualitative patterns replicate, with slower degradation rates than LLaMA-2.}
\label{tab:mistral_full}
\end{table}

Regularized-seed control produces a non-monotonic $\sigma$ trajectory with peak at generation~3 ($\sigma = 0.53$, declining to $0.34$). Three-condition filtering shows the same pattern: random filter indistinguishable from no filter ($p = 0.68$); quality filter sustains compositionality ($\sigma = 0.47$ at generation~10). The P4 compositionality peak occurs at the same generation (3) as LLaMA-2, suggesting the peak timing is architecture-independent and depends primarily on the bottleneck parameters. Cross-linguistic morphological regularization follows the same trajectory ordering (Turkish $>$ English $>$ German) with slightly lower absolute values at each generation. Complete per-generation tables for all metrics and conditions are provided in the released data.

\section{Cross-Linguistic Morphology Details}
\label{app:crossling}

German experiments use 150 verbs (75 strong, 75 weak) selected to span frequency quartiles. Strong$\to$weak regularization is measured as the proportion of strong verbs producing weak past tense forms (\emph{buk} $\to$ \emph{*backte}). Turkish experiments use 120 verbs testing vowel harmony adherence and suffix regularity. UniMorph paradigm tables \citep{coltekin2022measuring} provide the gold standard.

\section{Bias--Prediction Mapping}
\label{app:bias_mapping}

Following \citet{ferdinand2019regularization}'s demonstration that different bias types produce different evolutionary trajectories, we identify four specific bias--prediction mappings:

\begin{itemize}[nosep,leftmargin=*]
\item \emph{Frequency bias} from next-token prediction training: high-frequency tokens receive more gradient updates, creating an implicit frequency-dependent prior that drives P1 (frequency-dependent loss) and P5 (distributional narrowing).
\item \emph{Simplicity bias} from cross-entropy minimization: \citet{shah2020pitfalls} and \citet{kallini2024mission} showed LLMs favor regular morphological patterns, driving P2 (regularity increase).
\item \emph{Local coherence bias} from autoregressive factorization: \citet{mccoy2024embers} demonstrated that autoregressive training creates biases toward locally predictable sequences, driving P3's ordering (pragmatic constructions requiring non-local context are lost first).
\item \emph{Compression without grounding}: without communicative objectives, continued transmission converges to maximally compressible (degenerate) representations, driving P4 (non-monotonic compositionality).
\end{itemize}

These mappings allow us to derive specific predictions from the general iterated learning framework, moving beyond the generic ``inductive bias amplification'' characterization. Each mapping connects a well-documented LLM training property to a specific iterated learning dynamic, generating a testable prediction about which linguistic phenomena should change, in which direction, and at what relative rate.

\section{Full Zipf Exponent Data}
\label{app:zipf_full}

\begin{table}[h]
\centering
\small
\begin{tabular}{lcccc}
\toprule
\textbf{Gen.} & \multicolumn{2}{c}{\textbf{LLaMA-2}} & \multicolumn{2}{c}{\textbf{Mistral}} \\
\cmidrule(lr){2-3} \cmidrule(lr){4-5}
& $\alpha$ & $|V|$ & $\alpha$ & $|V|$ \\
\midrule
Human$^*$ & 1.08 & 31{,}204 & --- & --- \\
Seed & 1.07 & 28{,}412 & 1.07 & 28{,}412 \\
1 & 1.04 & 26{,}831 & 1.05 & 27{,}104 \\
3 & 0.98 & 23{,}247 & 1.01 & 24{,}619 \\
5 & 0.93 & 19{,}863 & 0.96 & 21{,}782 \\
7 & 0.88 & 16{,}541 & 0.91 & 18{,}905 \\
10 & 0.82 & 14{,}209 & 0.86 & 15{,}347 \\
\midrule
10\textsuperscript{QF} & 0.97 & 22{,}847 & 0.99 & 24{,}113 \\
\bottomrule
\end{tabular}
\caption{Zipf exponent ($\alpha$) and active vocabulary ($|V|$). Means over 5 runs. The power-law fit is preferred over lognormal ($\Delta$AIC $> 4$) and exponential ($\Delta$AIC $> 12$) alternatives at all generations \citep{clauset2009powerlaw}. \textsuperscript{QF}: quality filter.}
\label{tab:zipf}
\end{table}

\section{Human Regularization Overlay}
\label{app:human_overlay}

\begin{figure}[h]
	\centering
	\begin{tikzpicture}
		\begin{axis}[
			width=\columnwidth,
			height=5cm,
			xlabel={$\log_{10}$(Token Frequency)},
			ylabel={Regularity Preference (\%)},
			legend style={font=\scriptsize, at={(0.98,0.98)}, anchor=north east, fill=none, draw=none},
			xmin=1, xmax=6,
			ymin=45, ymax=100,
			grid=major,
			grid style={gray!20},
			]
			\addplot[only marks, mark=*, mark size=2pt, color=blue!70!black] coordinates {
				(1.5,92) (2.0,88) (2.5,82) (3.0,76) (3.5,71) (4.0,67) (4.5,64) (5.0,62) (5.5,61)
			};
			\addlegendentry{LLM Gen.~10}
			\addplot[thick, color=blue!70!black, domain=1:6, samples=50] {100 - 8*x};
			\addlegendentry{LLM log-linear fit}
			\addplot[only marks, mark=triangle*, mark size=2pt, color=red!70!black] coordinates {
				(1.5,89) (2.0,84) (2.5,79) (3.0,74) (3.5,70) (4.0,66) (4.5,63) (5.0,60) (5.5,58)
			};
			\addlegendentry{Human (M\&L 2016)}
			\addplot[thick, color=red!70!black, dashed, domain=1:6, samples=50] {95.5 - 7.2*x};
			\addlegendentry{Human log-linear fit}
		\end{axis}
	\end{tikzpicture}
	\caption{Frequency-dependent regularization gradient: LLM (blue) vs.\ human data from \citet{morgan2016frequency} (red). Both follow log-linear relationships with similar slopes ($-8.0$ vs.\ $-7.2$), consistent with a shared computational mechanism (frequency-independent bias + transmission bottleneck). $R^2 = 0.94$ for the LLM fit.}
	\label{fig:human_comparison}
\end{figure}
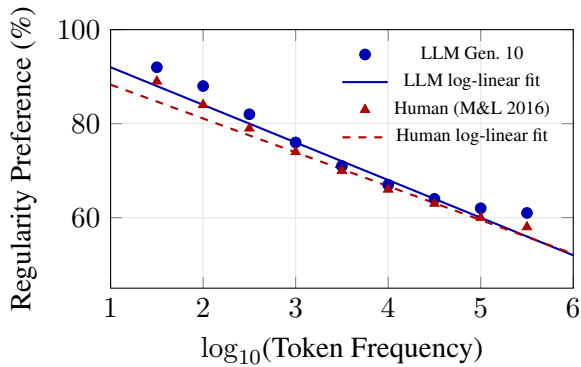

\section{Discussion: Construction Grammar and Human Parallels}
\label{app:cxg_discussion}

The LLM--human match ($R^2 = 0.94$ against the \citealp{morgan2016frequency} curve) suggests a shared mechanism: frequency-independent simplicity bias interacting with a transmission bottleneck. Whether LLMs share functional biases with humans remains open \citep{mahowald2024dissociating}. The generation-10 ``prior'' (dominant SVO, regular morphology, reduced construction diversity, sub-1.0 Zipf exponents) resembles adult learner biases \citep{hudson2005regularizing, hudson2009getting}; whether this more closely resembles children's \citep{culbertson2014language} or adults' learning remains open, though \citet{warstadt2022what} showed LLMs acquire syntax differently from children.

The P3 entrenchment correlation ($r = 0.73$) independently tests CxG predictions about productivity and resistance to loss \citep{goldberg2006constructions, bybee2010language}. \citet{misra2024language} showed LLMs learn rare constructions via transfer; our results show the converse: degradation through iterated learning makes rare constructions vulnerable. The eRank analysis confirms genuine structural erosion (eRank: $8.3 \to 4.1$).

\section{Extended Related Work}
\label{app:extended_related}

\paragraph{Compositionality and inductive biases.} Benchmarks: SCAN \citep{lake2018generalization}, COGS \citep{kim2020cogs}; meta-learning \citep{lake2023human}; evaluation frameworks \citep{hupkes2020compositionality, keysers2020measuring}. Emergent communication: \citet{lazaridou2017multiagent, lazaridou2020emergent} on language emergence; \citet{ren2020compositional} on compositionality from iterated learning; \citet{chaabouni2020compositionality} on compositionality--generalization; \citet{mu2021emergent} on emergence conditions. Inductive biases: architecture effects \citep{mccoy2020does}; LLM vs.\ child learning \citep{warstadt2022what}; autoregressive biases \citep{mccoy2024embers}; Bayesian prior distillation \citep{mccoy2025bayesian}; morphological biases \citep{kallini2024mission}; rare construction transfer \citep{misra2024language}; developmental modeling \citep{nikolaus2023modeling}; decoding distributions \citep{holtzman2020curious, meister2023locally}; predictability effects \citep{pimentel2024evidence}.

\end{document}